\documentclass[10pt,twocolumn,letterpaper]{article}
\pdfoutput=1

\usepackage{iccv}
\usepackage{times}
\usepackage{graphicx}
\usepackage{caption}
\usepackage{subcaption}
\usepackage{xcolor}
\usepackage{epsfig}
\usepackage{ dsfont }
\usepackage{amsmath}
\usepackage{amssymb}
\usepackage{booktabs}
\usepackage{stmaryrd}

\usepackage[pagebackref=true,breaklinks=true,letterpaper=true,colorlinks,bookmarks=false]{hyperref}

\iccvfinalcopy % *** Uncomment this line for the final submission

\def\iccvPaperID{6603} % *** Enter the ICCV Paper ID here

\ificcvfinal\fi
\begin{document}

%%%%%%%%% TITLE
\title{Multi-Features Guidance Network for partial-to-partial point cloud registration}
% pose extraction / pose understanding.... or maybe we should directly write PointNet based ....

%\author{
%	Hongyuan Wang \\
%	\texttt{fountainhy@163.com} \\
%	\And
%	Xiang Liu \\
%	\texttt{19B921006@stu.hit.edu.cn} \\
%	\And
%	Wen Kang \\
%	\texttt{} \\
%	\And
%	Zhiqiang Yan\\
%	\texttt{} \\
%	\And
%	Bingwen Wang \\
%	\texttt{} \\
%}

\author{\text{Hongyuan Wang}\\
	\texttt{fountainhy@163.com} \\
\and
\text{Xiang Liu}\\
 \texttt{19B921006@stu.hit.edu.cn} \\
\and
\text{Wen Kang}\\
\texttt{19B921007@stu.hit.edu.cn} \\
\and
\text{Zhiqiang Yan}\\
\texttt{18B921006@stu.hit.edu.cn} \\
\and
\text{Bingwen Wang}\\
\texttt{17B921012@stu.hit.edu.cn} \\
\and
\text{Qianhao Ning}\\
\texttt{19B921012@stu.hit.edu.cn} \\
%\and
%\text{Hou Jin}\\
%\texttt{19B921006@stu.hit.edu.cn} \\
%\and
%$^\text{1}$Carnegie Mellon University
%\and
%$^\text{2}$Fujitsu Laboratories Ltd.
%\and
%$^\text{3}$Argo AI.
%\and
%$^\text{4}$Apple.\\
%\and
%{\tt\small \{vsarode, xueqianl\}@andrew.cmu.edu}
%\and
%{\tt\small aoki-yasuhiro@fujitsu.com}
%\and
%{\tt\small \{hgoforth, slucey, choset\}@cs.cmu.edu}
%\and
%{\tt\small aruns@apple.com}
%% {\tt\small secondauthor@i2.org}
}

\maketitle
%\thispagestyle{empty}

%%%%%%%%% ABSTRACT
\begin{abstract}
 To eliminate the problems of large dimensional differences, big semantic gap, and mutual interference caused by hybrid features, in this paper, we propose a novel Multi-Features Guidance Network for partial-to-partial point cloud registration(MFG). The proposed network mainly includes four parts: keypoints' feature extraction, correspondences searching, correspondences credibility computation, and SVD, among which correspondences searching and correspondence credibility computation are the cores of the network. Unlike the previous work, we utilize the shape features and the spatial coordinates to guide correspondences search independently and fusing the matching results to obtain the final matching matrix. In the correspondences credibility computation module, based on the conflicted relationship between the features matching matrix and the coordinates matching matrix, we score the reliability for each correspondence, which can reduce the impact of mismatched or non-matched points. Experimental results show that our network outperforms the current state-of-the-art while maintaining computational efficiency. 
% Code is available at \href{https://github.com/vinits5/pcrnet.git}{https://github.com/vinits5/pcrnet}
\end{abstract}

%\begin{figure}[t!]
%    \centering
%    \includegraphics[width=\columnwidth]{figs/first_fig.pdf}
%    \caption{Comparison of different registration methods based on their robustness to noise and computation time with respect to object specificity. The iterative version of PCRNet exploits object specificity to produce accurate results. The PCRNet without iterations is computationally faster but compromises a little on accuracy. PointNetLK~\cite{aoki2019pointnetlk} exhibits good generalizability, but is not robust to noise. ICP~\cite{INTRO:ICP} is object-shape agnostic and slow for large point clouds, while Go-ICP~\cite{RW:GO_ICP} is computationally expensive.}
%    \label{fig:firstfig}
%\end{figure}

%%%%%%%%% BODY TEXT
\section{Introduction}
With the rapid development of sensor technology, there are more and more diversified methods to obtain 3d point cloud of the target scene (e.g., LiDAR, RGB-D camera, Stereo-camera, etc.), and 3D point cloud data become ubiquitous. Due to the point clouds collected from different views and distance, it is necessary to align two or more point clouds by estimating the rigid transformation between them, which play a vital role in 3d reconstruction, autonomous robot positioning, pose estimation, and other applications.

Sample Consensus Initial Alignment(SAC-IA) and Iterative Closest Point(ICP)~\cite{ICP} are widely used to solve the rigid transformation between point clouds. SAC-IA uses FPFH~\cite{FPFH} features to search for point correspondences, eliminates unreliable pairwise using RANSAC~\cite{RANSAC}, and finally compute the rigid transformation between point clouds. SAC-IA uses the informative feature descriptors to guide the matching points assignment, making the algorithm insensitive to point cloud initialization, whereas its performance is poorer than local registration algorithms. ICP iteratively assigns each point in one point cloud to the spatially closest point in another point cloud and computes the least-square rigid transformation between correspondences. However, owing to using spatial coordinates to guide correspondences search merely, ICP is significantly sensitive to initialization and outlier points, which hinders its use in practical applications. Therefore, many registration frameworks use a combination of feature descriptors and the point cloud's spatial position, using feature descriptors for a rough initial pre-alignment, followed by a more precise fine registration step based on the spatial position.

Benefiting from the rise of deep learning technology, researchers begin to study the use of data-driven point cloud processing methods. However, unlike traditional two-dimensional images, the point cloud is composed of a series of points, and the characteristics of unstructured make them problematic for use in deep learning architectures. PointNet~\cite{PointNet} is a pioneering framework applying deep learning techniques directly to unordered point cloud data. Inspired by PointNet, several deep learning architectures for point cloud registration tasks have been proposed and show that learning-based registration can be more robust and effective than traditional algorithms. Most of them~\cite{PointNetLK,DCP,PRNet,PCRNet,CorsNet} design a feature extraction network to compute local or global features and then regress correspondences and transformation between point clouds. Only a few methods~\cite{IDAM,RPM-Net} extract hybrid features to guide the correspondence assignment and transformation prediction, which improves registration performance to a certain extent. Nevertheless, using hybrid features also has several intrinsic flaws: 1) The dimension of spatial coordinates and geometric features varies greatly, and concatenate them directly is not conducive to exploiting point clouds' spatial position. 2) there are large semantic gaps between the spatial position and the shape features extracted from the feature extractor.3) The spatial coordinates and the shape features may interfere with each other and cannot give full play to their respective advantages.

To solve these problems mentioned above, and inspired by the combination of the coarse pre-alignment stage and the refinement stage, we propose the Multi-Features Guidance Network(MFGNet) for point cloud registration, which uses spatial coordinates and local features to guide the correspondences search jointly. Unlike approaches using hybrid features, MFGNet contains two different matching matrix computing branches: the coordinates matching matrix computing branch and the feature matching matrix computing branch. These two branches learn to assign correct correspondence independently, and then we fuse these two matching matrices to obtain the final matching matrix, followed by a differentiable singular value decomposition(SVD) layer to extract the rigid transformation. The main contributions of this paper are as follows:

1) A novel correspondences search module is proposed in this paper, which utilizes the shape features and the spatial coordinates to compute the matching matrix independently and finally fusing the matching results to obtain the final matching matrix.

2) We propose to use the conflicted relationship between the coordinates matching matrix and the features matching matrix to score the reliability for correspondences, which reduce the influence of mismatched or non-matched points.

3) Extensive analysis and tests of MFGNet are present, and experiments show that it achieves state-of-the-art performance on the ModelNet40 dataset.

4) The code and the pre-trained weight of MFGNet will be released to facilitate reproducibility and future research.
 
\begin{figure*}[t!]
	\centering
	\includegraphics[width=1\linewidth]{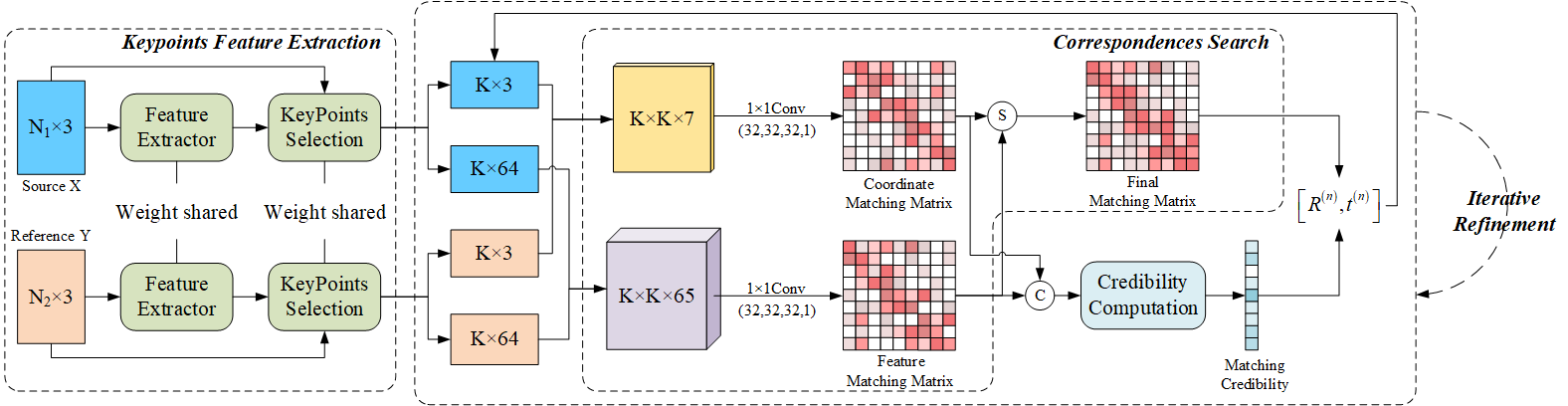}
	\caption{MFGNet overall Architecture.
	}
	\label{fig:overall_architecture}
\end{figure*}

%%%%%%%%%%%%%%%%%%%%%%%%%%%%%%%%%%%%%%%%%%%%%%%%%%%%%%%%%%%%%%%%%%%%%%%%%%%%%%%%%%%%
\section{Related Work}
\subsection{Traditional registration methods}
Over the last decades, numerous algorithms for point cloud registration have been proposed, which can be divided into two categories: coarse registration and fine registration. The coarse registration algorithm can generally be divided into three steps: 1) extract the structural features of points; 2) search the correspondences; 3) calculate the optimal transformation matrix. Among them, structural features extracting is vital for this type of method. Generally speaking, feature descriptors map neighboring points' features into histograms according to the number of neighboring points~\cite{feature1}, Euclidean distance~\cite{PFH}, the difference of normal~\cite{FPFH}, surface curvature~\cite{feature2}, etc.

Among the fine registration methods, ICP~\cite{ICP} is the most representative algorithm, which assigns each point in one point cloud to the spatially closest point in another point cloud and computes the least-square rigid transformation between correspondences iteratively. However, ICP is extremely sensitive to the initialization and outliers, and it is easy to fall into a local minimum. Therefore, many variants attempt to optimize the standard ICP by aggregating sensor uncertainty~\cite{G-ICP}, using other optimization~\cite{ICPVariant1}, or weighting matching points~\cite{ICPVariant2}, but most variants still strongly depend on the initialization. Therefore, many registration frameworks combine coarse registration with fine registration to avoid converging into local optimal.
\subsection{Learning-based registration methods}
Motivated by the excellent performance of deep learning in image processing, people begin to study learning-based point cloud registration methods. PointNetLK~\cite{PointNetLK} utilizes PointNet~\cite{PointNet} to extract global features for point clouds and then uses the modified Lucas-Kanade~\cite{Lucas-Kanade} algorithm to optimizes the transformation iteratively. PCRNet~\cite{PCRNet} compares the features extracted from PointNet to find the transformation between point clouds.  DCP~\cite{DCP} improves the initial features extracted from DGCNN~\cite{DGCNN} by incorporating the local and global information, enhancing the ability to approximate combinatorial matching. PRNet~\cite{PRNet} detects keypoints by comparing the L2 norms of features and uses Gumbel-Softmax~\cite{Gumbel-Softmax} with a straight-through gradient estimator to sample keypoints correspondence, which tackles a general partial-to-partial registration problem. IDAM~\cite{IDAM} utilizes hybrid features to search correspondences, which incorporates the shape features information and the spatial position information. Furthermore, a mutual-supervision loss is proposed to train the two-stage point elimination without extra annotations. Our work is similar to IDAM, but we independently use shape features and spatial coordinates to guide the correspondences search and fuse them finally. Besides, we calculate the credibility of matching points based on the matching results. RPM-Net~\cite{RPM-Net} uses the differentiable Sinkhorn~\cite{Sinkhorn} layer and annealing to get soft assignment matrix from hybrid features, which solves the problem of big difference in dimensions between spatial coordinates and local geometry to some extent, but it requires additional normal information. CorsNet~\cite{CorsNet} concatenates the local features with the global features and regresses correspondences between point clouds, followed by SVD to estimate the rigid transformation.

\section{Problem Statement}
The registration between two rigid point clouds refers to the task of finding a transformation to align different point clouds into one coordinate system. Given a source point cloud $\left\{X: x_{i} \mid i=1, \ldots N_{1}\right\} \subset \mathbb{R}^{3}$ and a target point cloud  $ \left\{Y: y_{j} \mid j=1, \ldots N_{2}\right\} \subset \mathbb{R}^{3} $, our objective is to find rigid transformation  $ \left \{ R,t \right \}  $ to align two point clouds, where  $ R \in S O(3) $ is a rotation matrix and $ t \in \mathbb{R}^{3} $ is a translation vector. Different from the most learning-based registration approach, the one-to-one correspondence between points is not required in our method, and $ X Y $ can cover different extents, which means $ N_{1} \neq N_{2} $in most cases. In this situation, we define a matching matrix $ \left\{M: m_{i j} \mid i=1, \ldots N_{1} ; j=1, \ldots N_{2}\right\} $ to represent whether $ x_{i}  $  corresponds to $ y_{j}  $ . Our objective is to find $ \left \{ R,t \right \}  $ to minimize the mapping error between correspondences :
\begin{align}
	\label{eq:Objective}
	\textbf{$ \min _{R, t} \sum_{j=1}^{N_{2}} \sum_{i=1}^{N_{1}} m_{i j}\left\|R x_{i}+t-y_{j}\right\|_{2}^{2} $}.
\end{align}
Where $ m_{i j} \in\{0,1\} $ is the $ i_{th}  $ row and $ j_{th} $ column element of the matching matrix $ M $, $ m_{ij} = 1 $ if point $ x_{i} $ corresponds to $ y_{j} $ and  $ m_{ij}=0 $ otherwise.

\section{Method}
In this section, we describe the proposed MFGNet point cloud registration model. Fig.~\ref{fig:overall_architecture} shows the architecture of MFGNet in detail. The model mainly consists of four components: 1) Keypoints' Feature Extraction, 2) Correspondences search, 3) Correspondences Credibility Computation, 4) SVD. The details of each component are described in the following sections.

\subsection{Keypoints' Feature Extraction}
The first stage is to extract the local feature descriptor from raw point clouds, which is essential to the subsequent correspondences search. A discriminative feature can well represent the local geometric structure and can find its correspondence easily. MFGNet can be compatible with both hand-crafted~\cite{FPFH,PFH,RoPS} and learning-based~\cite{PointNet,MMCNN,DGCNN} point cloud feature extraction methods, but this paper only uses the GNN network feature extraction layer to extract point cloud features. In our network, the dimension of the output features is 64.

\begin{figure*}[t!]
	\centering
	\includegraphics[width=0.9\linewidth]{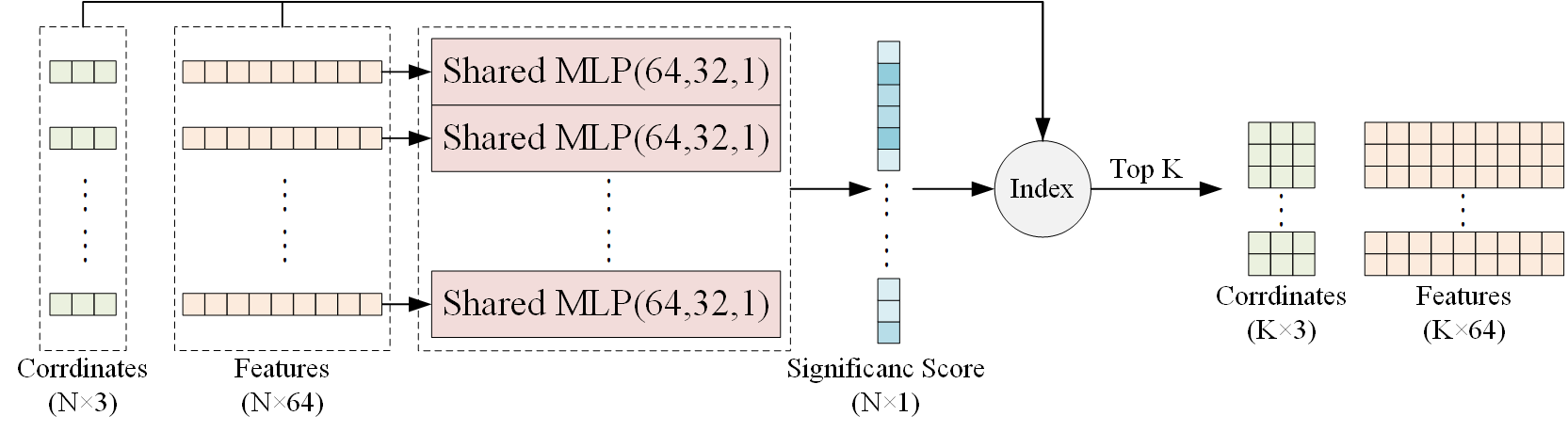}
	\caption{Network architecture for the keypoints selection module.
	}
	\label{fig:KeyPoints_Selection}
\end{figure*}

It's excessive-memory-required and time-consuming to use all points for correspondences search. Meanwhile, undiscriminating features will also lead to ambiguous matching results, which increase the difficulty of correspondence assignment. To avoid these problems mentioned above, we introduce the keypoints selection module to preserve discriminative features and eliminate undiscriminating features~\cite{IDAM}. Fig.~\ref{fig:KeyPoints_Selection} illustrates the pipeline of the keypoints selection module. Given the extracted features for each point, a multi-layer perceptron is applied to feature descriptors, and the significance score of features is obtained. A high significance score means a more discriminative feature, which is beneficial to the matching point search. We preserve the $ K $ points with the highest significance scores and discard the remaining points. We denote the keypoints preserved in the source and target point cloud as $ P_{S} $ and $ P_{T} $ and denote corresponding features of them as $ F_{S} $  and $ F_{T} $ . All of them will be used to guide the subsequent correspondence assignment.

\subsection{Correspondences search}
In this section, we elaborate on our correspondence matching strategy. Most of the existing learning-based methods merely use features extracted by MLP to guide correspondences search. However, many point cloud structures are repetitive, and it is challenging to obtain correct matching points without considering location information. To alleviating this problem, some networks use hybrid features learning from both shape feature and spatial coordinates to guide matching point assignment, but stack the features with different dimensions and semantics directly, making them interfere with each other and hindering the network from making full use of the advantages of different information. Inspired by the combined registration frameworks of coarse registration and fine registration, we argue that the shape features and the spatial coordinates can guide correspondences search independently, and fusing them adaptively at different registration stages can improve the registration accuracy.

We propose the correspondences search strategy guided jointly by shape features and spatial coordinates based on the above intuition. We obtain the keypoints' spatial coordinates and corresponding shape features from the point cloud in the last section. Given keypoints' shape features $ f_{S}(i) \in F_{S} $ and $ f_{T}(j) \in F_{T} $ for $ p_{S}(i) \in P_{S} $ and $ p_{T}(j) \in P_{T} $ , then we can form shape feature tensor and spatial coordinate tensor at iteration $ n $ as

\begin{align}
	\label{eq:feature_tensor}
	\begin{split}
	f^{(n)}(i, j)=\left[\left\|f_{S}(i)-f_{T}(j)\right\| ; \frac{f_{S}(i)-f_{T}(j)}{\left\|f_{S}(i)-f_{T}(j)\right\|}\right].
	\end{split}
%	\textbf{$ f^{(n)}(i, j)=\left[\left\|f_{S}(i)-f_{T}(j)\right\| ; \frac{f_{S}(i)-f_{T}(j)}{\left\|f_{S}(i)-f_{T}(j)\right\|}\right] \in F^{(n)} $}.
\end{align}
\begin{align}
	\label{eq:coordinate_tensor}
	\begin{split}
	s^{(n)}(i, j)=\left[\left\|p_{S}(i)-p_{T}(j)\right\| ; \frac{p_{S}(i)-p_{T}(j)}{\left\|p_{S}(i)-p_{T}(j)\right\|} ; p_{S}(i)\right]
	\end{split}
%	\textbf{$ s^{(n)}(i, j)=\left[\left\|p_{S}(i)-p_{T}(j)\right\| ; \frac{p_{S}(i)-p_{T}(j)}{\left\|p_{S}(i)-p_{T}(j)\right\|} ; p_{S}(i)\right] $}.
\end{align}
Where $ \left [ ; \right ]  $ means concatenate operation and $ \left \| \cdot  \right \|  $ denotes modulus of vectors. Unlike IDAM, which concatenates $ f_{S}(i) $ and $ f_{T}(j) $ directly, we use the distance between shape features to characterize the features' similarity. The shape feature tensor also contains the feature's relative direction, representing the difference between the features. The spatial coordinate tensor includes the distance, the direction between points, and the spatial coordinate of $ p_{S}(i) $ .

\begin{figure*}[t!]
	\begin{subfigure}{0.49\textwidth}
		\centering
		\includegraphics[width=0.9\linewidth]{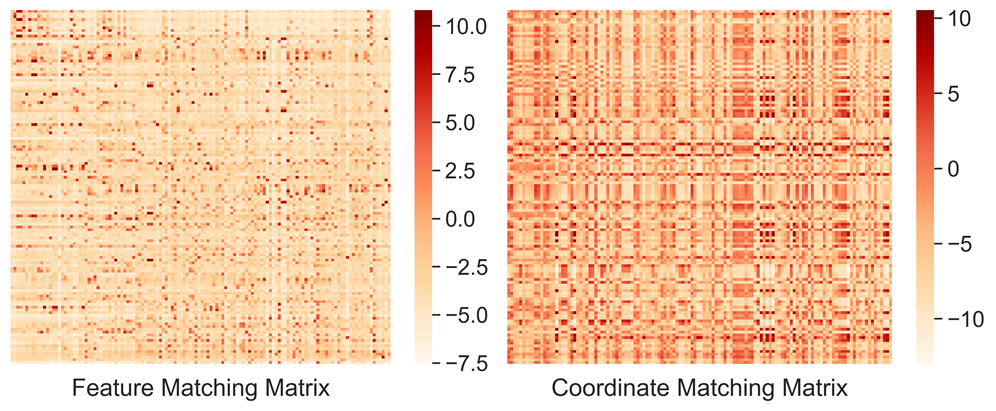}
		\caption{The Matching Matrix under the first iteration.}
		\label{fig:M0}
	\end{subfigure}
	~
	\begin{subfigure}{0.49\textwidth}
		\centering
		\includegraphics[width=0.9\linewidth]{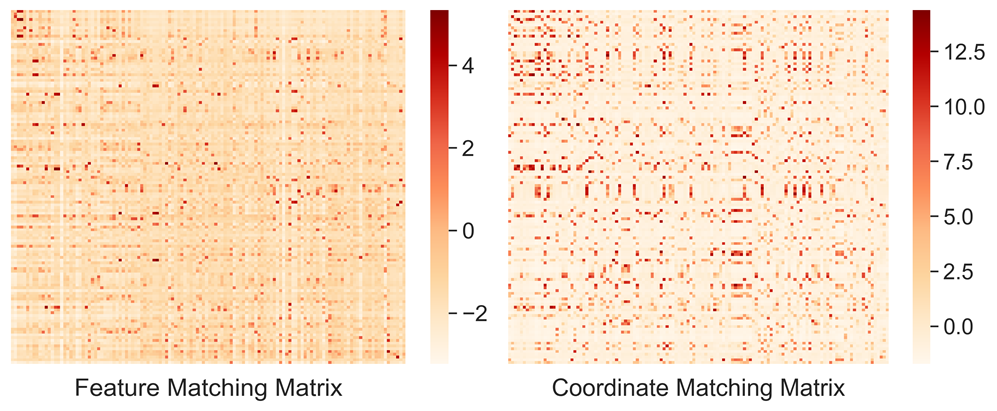}
		\caption{The Matching Matrix under the second iteration}
		\label{fig:M1}
	\end{subfigure}
	~
	\begin{subfigure}{0.49\textwidth}
		\centering
		\includegraphics[width=0.9\linewidth]{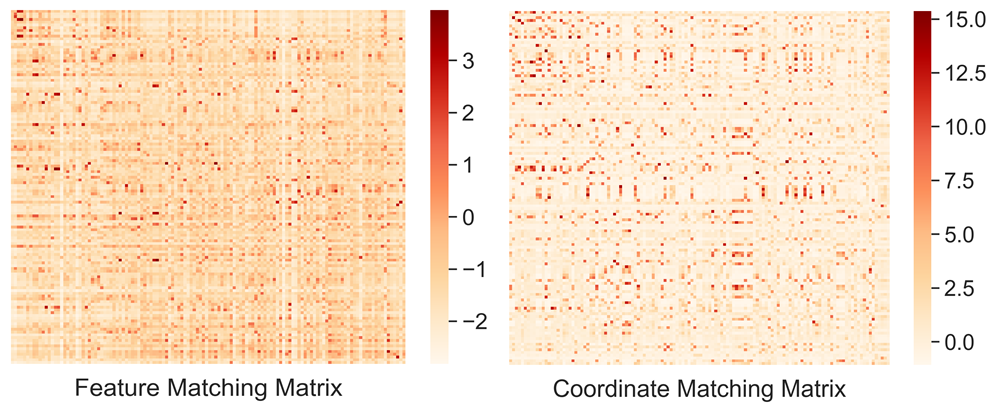}
		\caption{The Matching Matrix under the third iteration}
		\label{fig:M2}
	\end{subfigure}
	~
	\begin{subfigure}{0.49\textwidth}
		\centering
		\includegraphics[width=0.9\linewidth]{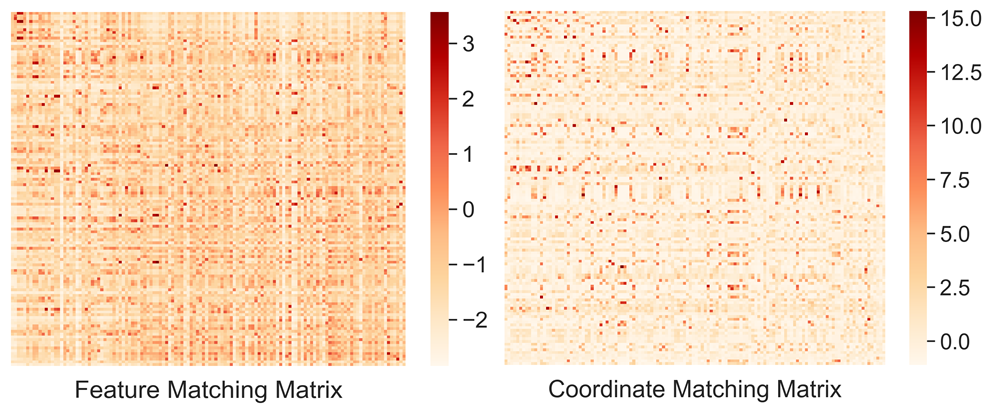}
		\caption{The Matching Matrix under the fourth iteration}
		\label{fig:M3}
	\end{subfigure}
	\caption{The Feature Matching Matrix and The Coordinate Matching Matrix results in different registration stages.}
	\label{fig:M_results}
\end{figure*}

In order to obtain the similarity of each point pair, a Multi-Layer Perceptron(MLP) is applied to the shape feature tensor and spatial coordinate tensor, respectively. In our network, we implement MLP with fully connected(FC) layers, each followed by ReLU activation and batch normation(BN) except the last layer. It is noted that the two MLPs are not weight shared, which allows them to guide the correspondence matching independently. The two MLPs output a coordinate matching matrix and a feature matching matrix respectively, each of which can be regarded as the correspondence matching result between point clouds, and a high response means a high matching probability, while a low response value means that the point pair is unlikely to be a correspondence. Logically, we add the two matrices to obtain the final matching matrix M, which enhances the response of the correct match point and suppresses the wrong match result. Obviously, different matching matrices have different credibility in different registration stages. Our network can learn this well and give them appropriate response values, which can be verified in fig.~\ref{fig:M_results}. We can see that the highest response ratio between the feature matching matrix and the coordinate matching matrix is approximately 1:1, 1:3, 1:4, and 1:5 in the four iterations.

\subsection{Correspondences Credibility Computation}
The previous section adopted the correspondences search strategy guided jointly by shape features and spatial coordinates, generating a shape feature matching matrix and a spatial coordinate matching matrix. In this section, these two matching matrices will be used to calculate the matching results' credibility. Intuitively, when the matching result of shape features is consistent with the matching result of spatial coordinates, it is reasonable to believe the matching results are correct in high probability. In contrast, when two matching results conflict, the degree of conflict between them and the registration stage should be considered to judge whether the matching result is credible. Therefore, to reduce the impact of incorrect matching points, we introduce a credibility calculation module to evaluate matching results' reliability.

\begin{figure}[t!]
	\centering
	\includegraphics[width=0.9\linewidth]{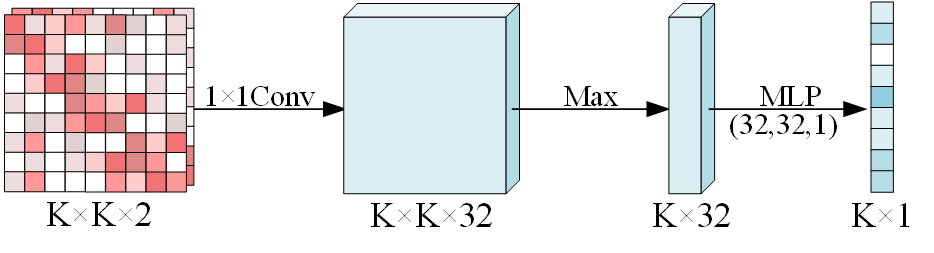}
	\caption{Network architecture for the correspondences credibility computation module.
	}
	\label{fig:Credibility}
\end{figure}

Fig.\ref{fig:Credibility} shows the specific structure of the correspondences credibility calculation module. To obtain the conflicted relationship between different matching matrices, we concatenate the shape feature matching matrix and the spatial coordinate matching matrix and map it to the high-dimensional feature space with a $ 1 \times 1 $ convolution operation. Then the element-wise max aggregation and MLP are used to obtain the matching results' credibility. A high score means high credibility and has a higher weight in the subsequent weighted SVD layer, while a low score has little or no influence in the subsequent transformation solution process. We define the weight of the $ i_{th} $ correspondence result as
\begin{align}
	\label{credibility}
	\omega_{i}=\frac{c(i) \cdot \mathds{1} \llbracket c(i) \geq \operatorname{median}(c) \rrbracket}{\left.\sum_{i} c(i) \cdot \mathds{1} \llbracket c(i) \geq \operatorname{median}(c)\right \rrbracket}
\end{align}
where $ c(i) $ denotes the credibility of the matching point, and $ \mathds{1} \llbracket \rrbracket $ is an indicator function to judge whether the matching result's credibility is greater than the median of scores. Once the correspondence assignment and its credibility are obtained, the final step is to estimate the rigid transformation $ {R,t} $. Like most existing methods, we use the weighted SVD to solve the transformation matrix, which can be defined as Eq.~\ref{eq:weighted_SVD}. The slight difference between Eq.~\ref{eq:Objective} and Eq.~\ref{eq:weighted_SVD} is that $ \omega_{i} $ is a decimal between $ 0 \sim 1 $ while $ m_{ij} \in \{0,1\} $.
\begin{align}
	\label{eq:weighted_SVD}
	R, t=\underset{R, t}{\operatorname{argmin}} \sum_{i} \omega_{i}\left\|R p_{S}(i)+t-p_{T}\left(j^{*}\right)\right\|
\end{align}

\begin{figure*}[t!]
	\begin{subfigure}{0.9\textwidth}
		\centering
		\label{fig:clean_results}
		\includegraphics[width=0.9\linewidth]{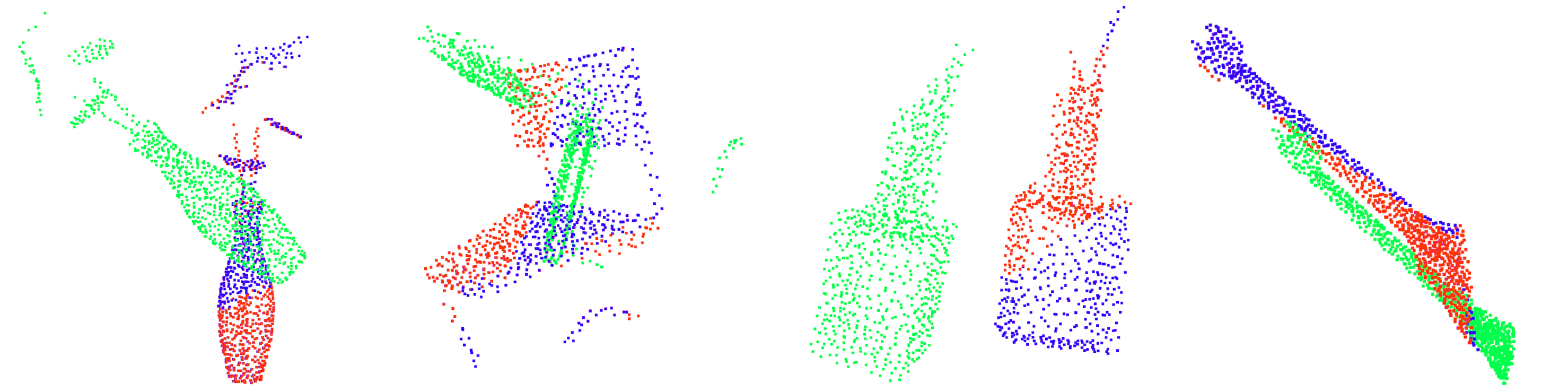}
		\caption{Qualitative registration expamles on unseen shapes}
	\end{subfigure}
	\begin{subfigure}{0.9\textwidth}
		\centering
		\label{fig:unseen_results}
		\includegraphics[width=0.9\linewidth]{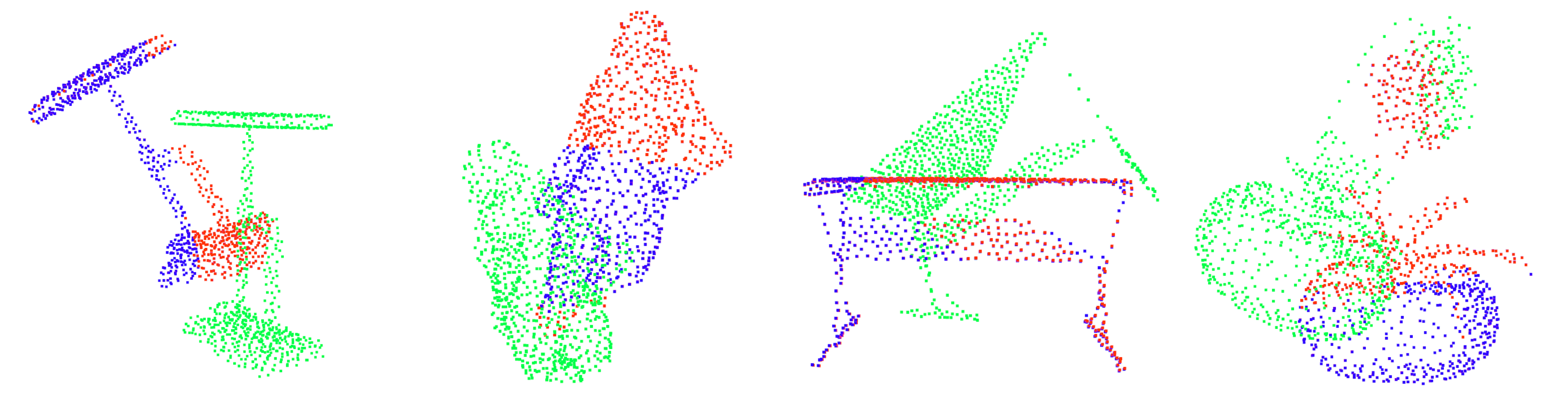}
		\caption{Qualitative registration expamles on unseen categories}
	\end{subfigure}
	\begin{subfigure}{0.9\textwidth}
		\centering
		\label{fig:gaussian_results}
		\includegraphics[width=0.9\linewidth]{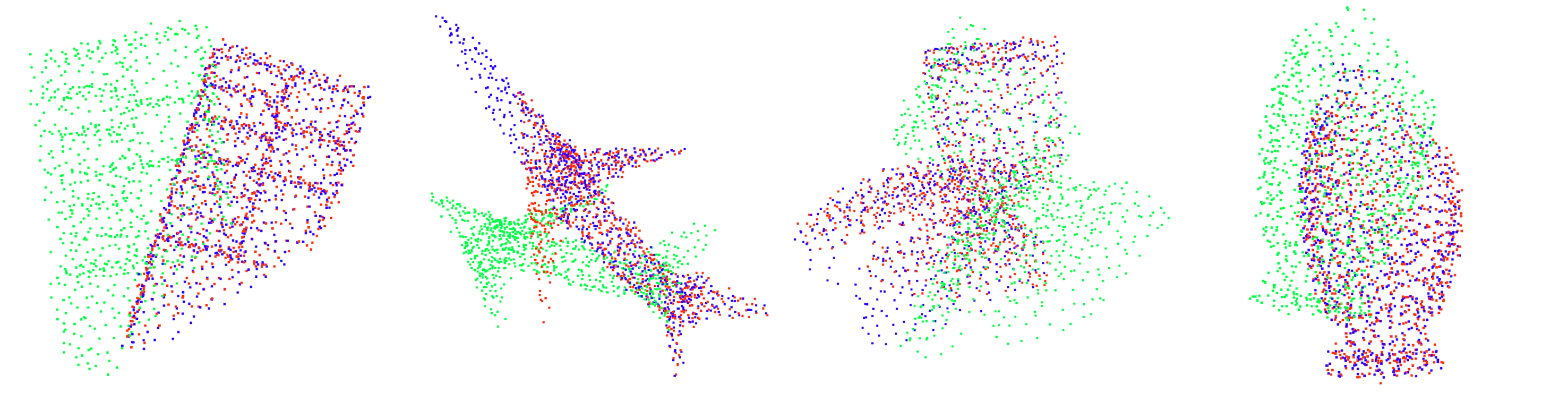}
		\caption{Qualitative registration expamles on noisy unseen shapes}
	\end{subfigure}
	\caption{Qualitative registration expamles under different experimental setups}
	\label{fg:qualitative_results}
\end{figure*}

\subsection{Loss Functions}
The loss function of our network consists of three parts: the keypoints selection loss $ L_{keypoint} $, the correspondences search loss $ L_{matching} $, and the correspondences credibility computation loss $ L_{credibility} $. Since we do not have access to keypoints annotation directly, we use Mutual-Supervision Loss~\cite{IDAM} to train our network. The principle behind it is that key points tend to have a low entropy because they are confident in matching. The loss of keypoints selection is defined as
\begin{align}
	\label{eq:keypoint_loss}
	L_{\text {keypoint }}=\frac{1}{K} \sum_{i=1}^{K}\left(s(i)-\sum_{j=1}^{K} M(i, j) \log (M(i, j))\right)^{2}
\end{align}
Where $ s(i) $ is the significance scores for $ p_{S}(i) $, and we only use the keypoints selection loss for the first iteration.

The loss of correspondence matching loss for the $ n_{th} $ iteration is defined as
\begin{align}
	\label{eq:matching_loss}
	L_{\text {matching}}^{(n)}=\frac{1}{K} \sum_{i=1}^{K}-\hat{y}_{i} \log \left(M^{(n)}\left(i, j^{*}\right)\right)
\end{align}
Where $ j^{\ast }  $ is the index of the point that is closest to $ p_{S}(i) $ under ground truth transformation, and $ \hat{y } _{i} $ is the label to judge whether the distance between $ p_{S}(i) $ and $ p_{T}(j^{\ast}) $ is less than the distance threshold $ th $.

The loss of correspondences credibility computation for the $ n_{th} $ iteration is defined as
\begin{align}
	\label{eq:credibility_loss}
	L_{\text {credibility }}^{(n)}=\frac{1}{K} \sum_{i=1}^{K}-\hat{m}_{i} \log (c(i))-\left(1-\hat{m}_{i}\right) \log (1-c(i))
\end{align}
Where $ \hat{m}_{i} $ is the label whether the correspondence distance under ground truth transformation is less than the distance threshold $ th $.

The overall loss is the weighted sum of the three losses
\begin{align}
	\label{eq:total_loss}
	L_{\text {total}}=\sum_{n}\left(\lambda(n) L_{\text {keypoint }}+L_{\text {matching }}^{(n)}+L_{\text {credibility }}^{(n)}\right)
\end{align}
Where
\begin{align}
	\label{eq:weight_loss}
	\lambda(n)=\left\{\begin{array}{ll}
		1 & n=1 \\
		0 & n \neq 1
	\end{array}\right.
\end{align}

\setlength{\tabcolsep}{4mm}
\begin{table*}[t!]
	\centering
	\caption[m1]{Results for testing on unseen point clouds}
	\label{tb:clean_data}
	\begin{tabular}{c c c c c}
		\hline
		Model & RMSE(R) & MAE(R) & RMSE(t) & MAE(t)\\
		\hline
		ICP & 34.99 & 25.46 & 0.296 & 0.253\\
		FGR & 8.72 & 1.86 & 0.017 & 0.005\\
		SAC-IA & 6.04 & 1.90 & 0.016 & 0.012\\
		SAC-IA+ICP & 6.64 & 0.71 & 0.022 & 0.005\\
		PointNetLK & 16.74 & 7.55 & 0.045 & 0.025\\
		DCP & 6.71 & 4.45 & 0.027 & 0.020\\
		PRNet & 3.20 & 1.45 & 0.016 & 0.010\\
		IDAM & 2.95 & 0.76 & 0.021 & 0.005\\
		\hline
		MFGNet & $ \bf{1.56} $ & $ \bf{0.39} $ & $ \bf{0.006} $ & $ \bf{0.002} $\\
		\hline
	\end{tabular}
\end{table*}

\section{Experiment}

\subsection{Dataset and Experimental Setup}
We evaluate our model on the ModelNet40 dataset, which is widely used to train point cloud registration networks. ModelNet40~\cite{ModelNet40} contains 12311 CAD models from 40 human-made categories, split officially into two parts: the training set containing 9843 models and the testing set containing 2468 models. We use the processed data from~\cite{PointNet}, which sample 2048 points randomly from the models and normalize them into a unit sphere. For consistency, a point cloud with 1024 points is sampled, and a rigid transformation is randomly generated (rotations within [0,45] along each axis and translation within [-0.5, 0.5]). Following~\cite{PRNet,IDAM}, we fix a random point and preserve 768 points closest to this point for each point cloud and generate partially overlap point clouds.

In the experiment, the top 128 points are preserved as key points, the distance threshold th is set to 0.05, and the number of iterations is set to 4. We randomly sample 64 "positive" points that have correspondence in the target point cloud and 64 "negative" points that do not have correspondence during training. We train our model using Adam~\cite{Adam} optimizer for 50 epochs, with an initial learning rate of 0.0001 and a weight decay of 0.001. We divide the learning rate by 10 at 40 epochs.

We compare our model to classic registration methods(including SAC-IA~\cite{FPFH}, ICP~\cite{ICP}, FGR~\cite{FGR},SAC-IA+ICP) and recently learning-based registration methods(including PointNetLK~\cite{PointNetLK}, DCP~\cite{DCP}, PRNet~\cite{PRNet},IDAM~\cite{IDAM}). All the learning-based methods are trained on the same training set, and we provide the root mean square error(RMSE) and the mean absolute errors(MAE) for rotation matrix and translation vectors to evaluate all methods. For the rotation matrix, the unit of mean absolute error is the degree.
\subsection{Experimental Results}
\subsubsection{Full Dataset Train \& Test}
In the first experiment, we evaluate our model on the full official training/testing sets. Both the training set and test set contain point clouds from all the 40 categories. We list the quantitative results for all methods in Table~\ref{tb:clean_data}. As we can see from the results, our method outperforms all the other registration algorithms, both traditional and learning-based methods.

We can see that the ICP algorithm performs poorest due to the lack of a good initial position, which means that only using spatial coordinates to guide the registration is insufficient. FGR and SAC-IA obtain better results for using shape features to guide registration. Besides, SAC-IA+ICP, which utilizes both local shape features and spatial coordinates, performs best in traditional methods, even better than learning-based methods. Inspired by SAC-IA+ICP, we design a new network called MFGNet, which takes full advantage of point clouds' spatial coordinates and shape features. Experimental results show that MFGNet is superior to other methods in all indicators.
\subsubsection{Category Split}
To test the generalizability of different methods, we conducted the second experiment, which trains models on the first 20 categories in the training set while evaluating them on the other 20 categories in the test set. The quantitative results are summarized in Table ~\ref{tb:unseen_category}.
\setlength{\tabcolsep}{4mm}
\begin{table*}[t!]
	\centering
	\caption[m1]{Results for testing on unseen categories}
	\label{tb:unseen_category}
	\begin{tabular}{c c c c c}
		\hline
		Model & RMSE(R) & MAE(R) & RMSE(t) & MAE(t)\\
		\hline
		ICP & 34.27 & 25.63 & 0.294 & 0.250\\
		FGR & 8.32 & 1.21 & 0.013 & 0.004\\
		SAC-IA & 3.89 & 1.71 & 0.019 & 0.013\\
		SAC-IA+ICP & 6.01 & 0.52 & 0.014 & 0.004\\
		PointNetLK & 22.94 & 9.66 & 0.061 & 0.033\\
		DCP & 9.77 & 6.95 & 0.034 & 0.025\\
		PRNet & 4.99 & 2.33 & 0.021 & 0.015\\
		IDAM & 3.42 & 0.93 & 0.022 & 0.005\\
		\hline
		MFGNet & $ \bf{1.476} $ & $ \bf{0.43} $ & $ \bf{0.008} $ & $ \bf{0.002} $\\
		\hline
	\end{tabular}
\end{table*}
\setlength{\tabcolsep}{4mm}
\begin{table*}[t!]
	\centering
	\caption[m1]{Results for testing on point clouds with Gaussian noise}
	\label{tb:gaussian_noise}
	\begin{tabular}{c c c c c}
		\hline
		Model & RMSE(R) & MAE(R) & RMSE(t) & MAE(t)\\
		\hline
		ICP & 33.86 & 25.07 & 0.292 & 0.249\\
		FGR & 27.13 & 12.06 & 0.064 & 0.036\\
		SAC-IA & 11.95 & 3.32 & 0.027 & 0.016\\
		SAC-IA+ICP & 12.53 & 2.44 & 0.022 & $ \bf{0.008} $\\
		PointNetLK & 19.94 & 9.08 & 0.057 & 0.032\\
		DCP & 6.88 & 4.53 & 0.028 & 0.021\\
		PRNet & 4.32 & 2.05 & $ \bf{0.017} $ & 0.012\\
		IDAM & 3.72 & 1.85 & 0.023 & 0.011\\
		\hline
		MFGNet & $ \bf{3.56} $ & $ \bf{1.52} $ & 0.019 & $ \bf{0.008} $\\
		\hline
	\end{tabular}
\end{table*}

Compared with Table ~\ref{tb:clean_data}, it can be found that the invisible categories have a little side effect on the traditional methods because they do not need to use the training set to learn parameters. However, all the learning-based methods' performance has a certain degree of degradation for lacking the prior knowledge of test categories. Nevertheless, benefiting from the combination of shape features and location features, MFGNet has less performance degradation and still outperforms all other methods.

\subsubsection{Gaussian Noise}

In practical applications, the point cloud inevitably contains outliers and noise. To verify the algorithm's effectiveness under noise, we conducted the last point cloud registration experiment by adding Gaussian noise. In this experiment, we add Gaussian noise with a standard deviation 0.01 to all point clouds and clip the noise to $ \left [ -0.5,0.5 \right ]  $. We follow the first experiment's other settings (full dateset train \& test) and summarize the quantitative results in Table~\ref{tb:gaussian_noise}. As shown in Table~\ref{tb:gaussian_noise}, since Gaussian noise has an immense impact on shape features and spatial coordinates, all registration methods' performance degrades greatly. Nevertheless, MFGNet has the best performance among all registration algorithms. We show the results of our model under different experimental setting in fig.~\ref{fg:qualitative_results}.

\subsubsection{Efficiency}

To compare the efficiency of different models, we calculate the average inference time for each model. We use the implementations of ICP, SAC-IA, FGR in Intel Open3D~\cite{Open3d}, and the official implementation of PointLK, DCP, PRNet, and IDAM released by the authors. In our experiment, we set the maximum number of ICP iterations to 2000, the maximum number of RANSAC iterations to 4×106, and the maximum number of validation steps to 500. The experiments are performed on a 3.0GHz Intel i7-9700 CPU and an Nvidia GeForce RTX 2080 SUPER GPU. We test the efficiency of the models on points with 512, 1024, and 2048 points, and use a batch size of 1 for all learning-based methods during testing. The results are summarized in Table~\ref{tb:efficiency}. It can be seen that MFGNet is significantly faster than SAC-IA+ICP, which has comparable performance as our method. At the same time, MFGNet have similar efficiency with other learning-based registration networks.
\setlength{\tabcolsep}{2mm}
\begin{table*}
	\centering
	\caption{Average inference time required for registering point clouds of various sizes(in milliseconds)}
	\label{tb:efficiency}
	\begin{tabular}{c c c c c c c c c c}
		\hline
		 & ICP & FGR & SAC-IA & SAC-IA+ICP & PointNetLK & DCP & PRNet & IDAM & MFGNet\\
		\hline
		 512 points & 8.3 & 14.1 & 92.1 & 86.9 & 118.5 & 13.2 & 83.4 & 34.0 & 44.9\\
		 1024 points & 10.7 & 27.8 & 96.7 & 98.1 & 120.3 & 22.5 & 134.3 & 34.7 & 45.8\\
		 2048 points & 16.5 & 54.6 & 120.5 & 149.6 & 124.6 & 56.25 & 219.1 & 36.6 & 47.3\\
		 \hline
	\end{tabular}
\end{table*}
\setlength{\tabcolsep}{4mm}
\begin{table*}
	\centering
	\caption{the quantitative results under different iteration times}
	\label{tb:ablation}
	\begin{tabular}{c c c c c c}
		\hline
		$ n $ & RMSE(R) & MAE(R) & RMSE(t) & MAE(t) & Inference time\\
		\hline
		2 & 3.99 & 1.99 & 0.023 & 0.012 & 31.8\\
		3 & 3.64 & 1.62 & 0.019 & 0.009 & 38.3\\
		4 & 3.56 & 1.52 & 0.019 & 0.008 & 45.8\\
		5 & 3.80 & 1.57 & 0.020 & 0.009 & 52.2\\
		6 & 3.67 & 1.46 & 0.024 & 0.009 & 59.6\\
		\hline
	\end{tabular}
\end{table*}

\subsubsection{Ablation Study}

In this section, we compare the performance of the MFGNet algorithm under different iteration times $ n $. We follow the third experiment settings (training point clouds with Gaussian noise) and set $ n $ to 2, 3, 4, 5, and 6, respectively. Table~\ref{tb:ablation} shows the quantitative results under different iteration times. It can be seen that when $ n=2 $, the MFGNet can achieve similar performance as PRNet but 3 times faster than PRNet. When $ n=3 $, the performance of MFGNet has already surpassed the IDAM. When $ n>4 $, there is no obvious improvement in performance. To balance the efficiency and the performance, we set $ n $ to 4 in all experiments conducted above.

\section{Conclusion}

In this paper, we propose a novel network for point cloud registration, called MFGNet. Inspired by the combination of coarse registration and fine registration, we propose a strategy that utilizes shape features and spatial coordinates to guide the correspondences search, enhancing the network's ability to find correct correspondences. We also proposed a correspondences credibility computation module based on the feature matching matrix and the coordinate matching matrix, which scores the correspondences' reliability and reduces the wrong matching results' side impact. The experimental results show that our algorithm achieves state-of-the-art performance on the ModelNet40 dataset under different experimental settings.

%Acknowledgement: Mention Tejas. 
\section*{Acknowledgement}
This preprint has not undergone peer review or any post-submission improvements or corrections. The Final Version of this article is published in 	\textit{
	Neural Computing and Applications}, and is available online at \href{https://doi.org/10.1007/s00521-021-06464-y}{https://doi.org/10.1007/s00521-021-06464-y}. 

{\small
\bibliographystyle{plain}
\bibliography{egbib}
}
%\bibliography{egbib.bib}
%\bibliographystyle{ieee}

\end{document}